\documentclass[a4paper,conference]{IEEEtran}
\usepackage{cite}

\ifCLASSINFOpdf
  \usepackage[pdftex]{graphicx}
\else
\fi
\usepackage{amsmath}
\usepackage[inline]{enumitem}
\usepackage[free-standing-units=true]{siunitx}

\begin{document}
%
\title{Visual Feature Encoding for GNNs on Road Networks}



%
\author{\IEEEauthorblockN{Oliver Stromann\IEEEauthorrefmark{1,2},
Alireza Razavi\IEEEauthorrefmark{2} and 
Michael Felsberg\IEEEauthorrefmark{1}}
\IEEEauthorblockA{\IEEEauthorrefmark{1}Computer Vision Laboratory, Linköping University, Sweden \\ Email: oliver.stromann@liu.se, michael.felsberg@liu.se}
\IEEEauthorblockA{\IEEEauthorrefmark{2}Autonomous Transport Solution Research, Scania CV AB, Sweden\\
Email: alireza.razavi@scania.com}}


\maketitle

\begin{abstract}
In this work, we present a novel approach to learning an encoding of visual features into graph neural networks with the application on road network data. We propose an architecture that combines state-of-the-art vision backbone networks with graph neural networks. 
More specifically, we perform a road type classification task on an Open Street Map road network through encoding of satellite imagery using various ResNet architectures. Our architecture further enables fine-tuning and a transfer-learning approach is evaluated by pretraining on the NWPU-RESISC45 image classification dataset for remote sensing and comparing them to purely ImageNet-pretrained ResNet models as visual feature encoders. 
The results show not only that the visual feature encoders are superior to low-level visual features, but also that the fine-tuning of the visual feature encoder to a general remote sensing dataset such as NWPU-RESISC45 can further improve the performance of a GNN on a machine learning task like road type classification.
\end{abstract}


\IEEEpeerreviewmaketitle

\section{Introduction}

Connected and autonomous vehicles are an emerging trend in the transport sector, and the road network is the core infrastructure on which these vehicles operate. Knowledge about road networks is the key for making good decisions for connected and autonomous vehicles; for immediate actions of individual traffic users, as well as large-scale traffic management in the transportation system as a whole. It is therefore of large importance to find meaningful and efficient representations of road networks to enable or simplify the knowledge generation~\cite{stromann2021learning}.
 
However, the structural information encoded in spatial graphs like road networks has two shortcomings: first, the incompleteness of the encoded information and second the absence of information that cannot easily be encoded in discretized variables. On the other hand, unstructured image data contains complementary information with the potential to fill in those gaps~\cite{stromann2021learning}. 

\subsection{Contributions}
In~\cite{stromann2021learning} we have shown how spatial graphs can be enriched with hand-crafted low-level visual features to improve performance of graph neural networks (GNNs) on a machine learning task. In the present work, we propose to utilize state-of-the-art pretrained and fine-tuned vision backbone networks as visual feature encoders for GNNs to fully exploit the image data that can be associated to a spatial graph.

We perform a road type classification task using a GNN over crowd-sourced road network data from OpenStreetMap (OSM)~\cite{OpenStreetMap} with visual feature encodings produced by ResNets~\cite{he2016resnet} using high-resolution satellite imagery from Maxar Technologies~\cite{maxar2021}. 

The results do not only show that the visual feature encoders are superior to hand-crafted low-level visual features, but also that a fine-tuning of the visual feature encoder to a general remote sensing dataset such as NWPU-RESISC45~\cite{cheng2017remote} can further improve the performance of a GNN on a machine learning task like road type classification.

\section{Related Work}
\subsection{Geographical Data on Road Networks}
\subsubsection{Road Network Graphs}
Road network graphs are spatial graphs, which are nowadays made available by mapping sources like Google Maps, Microsoft Bing Maps, or OSM (see~\cite{veenendaal2017review} and references therein). A road network graph is an attributed directed spatial graph, with intersections as nodes and roads connecting these intersections as edges~\cite{stromann2021learning}. Additional information about intersections and roads can be associated to the node and edge attributes. Intersection attributes can be information like the coordinates or the type of right-of-way; while road attributes can be features information about the number of lanes, the speed limit or the geometry of the road. 

The crowd-sourced road network graphs from OSM~\cite{OpenStreetMap} however, often have missing attributes or attributes that are inconsistently set~\cite{funke2015automatic, haklay2010good}. For example, information about the number of lanes is often limited to the case when it deviates from the norm - which might vary over different geographical regions. Moreover, the attributes might also be subject to changes over time.  Lastly, due to the data structure of road network graphs, usually only discretized information is stored in the attributes, i.e. each attribute can hold only one value, even if in reality the attribute varies over the length of the road.

\subsubsection{Remote Sensing Data}
A suitable complement to the road network graphs is remote sensing data. This is data collected from air or spaceborne sensors like radars, lidars or cameras. Due to the bird's-eye perspective, it has a complete ground coverage without missing data. Nowadays, analysis-ready remote sensing data  is available that has undergone  atmospheric, radiometric, and topographic error correction~\cite{maxar2021}. It is usually composited of temporal stacks of imagery, such that dynamic objects like clouds and vehicles are removed~\cite{dwyer2018analysis}.

Especially, with the rise of free and open platforms like Google Earth Engine, Open Data Cube or openEO, remote sensing data is becoming more accessible and can be conveniently included as additional sources of information (see~\cite{gomes2020overview} and references therein). However, in the context of road networks, aerial imagery of very high spatial resolution below 1 m is favourable~\cite{mnih2013aerial}, which is mostly still proprietary data.

\subsection{Machine Learning Concepts}
\subsubsection{Graph Neural Networks (GNNs)}
Recent years brought a surge of publications on GNNs. Many architectures have been proposed to produce deep embeddings of nodes, edges, or whole graphs~\cite{hamilton2020graph-rep,zhou2020graph}. With low-dimensional embeddings~\cite{grover2016node2vec,perozzi2014deepwalk}, convolutions~\cite{kipf2016semi}, skip-connections~\cite{xu2018representation}, dropout- and normalization-layers~\cite{chen2018stochastic}, or attention mechanisms~\cite{zhang2018gaan,velickovic2018graph}, many techniques from other deep learning domains such as computer vision and natural language processing were adapted to GNNs~\cite{zhou2020graph}.

\subsubsection{Vision Backbone Networks}
Over the past decade, deep learning in computer vision has made giant leaps, and many architectures have been proposed to improve performance on machine learning tasks such as image classification. At the core, many of these architectures roughly follow the structure of multiple consecutive convolutional layers, making up a convolutional neural network (CNN) followed by fully-connected (FC) layers. 

The CNN which extracts image encodings of the image input, can be regarded as \textit{vision backbone}, while the FC layers which map the image encodings to class labels, can be regarded as the \textit{classifier head}.

Models of such architectures pretrained on the extensive ImageNet~\cite{russakovsky2015imagenet} dataset can be adapted to novel datasets and tasks using transfer learning~\cite{hoeser2020object,pires2020cnnTransferRS}. This adaption can be done in two ways: by feature extraction or by fine-tuning~\cite{li2017learning}. In feature extraction, the \textit{backbone} is kept static, but the \textit{head} is further trained to the new task. In fine-tuning, the parameters of both the \textit{backbone} and \textit{head} are updated through further training. In both cases, the \textit{head} can be replaced by another one suited to the novel task. This way, the architecture can be changed to a different learning domain, such as regression.

In our work, we use ResNet architectures~\cite{he2016resnet} as backbones. ResNets use residual connections between the convolutional layers, mitigating the vanishing gradient problem and therefore enabling deeper stacks of convolutional layers. ResNets are frequently used in transfer learning approaches, achieving state-of-the-art results~\cite{hoeser2020object}.
 
\subsection{Machine Learning on Geographical Data}
\subsubsection{Machine Learning on Road Networks}
Application such as location-based services, fleet planning, next-delivery suggestions, or traffic flow optimization help to tackle congestions and sustainability issues in the growing urban transportation systems~\cite{veres2019deep}. Many of these applications are enabled or can be enhanced with machine learning on road networks. Typical machine learning tasks like classification, regression, sequence prediction or clustering on road networks are enabling vehicle-centric predictions such as next-turn, destination and time of arrival predictions or network-centric predictions such as speed limit, travel time or traffic flow predictions~\cite{stromann2021learning,wu2020hrnr,liu2020graphsage,jepsen2020relational,gharaee2021graph,he2020roadtagger}.

The spatial graph representations of a road network makes it a prime candidate for utilizing GNNs. Jepsen \textit{et al.}~\cite{jepsen2020relational} propose a relational fusion network (RFN), which use different representations of a road network concurrently to aggregate features. Wu \textit{et al.}~\cite{wu2020hrnr} developed a hierarchical road network representation (HRNR) in which a three-level neural architecture is constructed that encodes functional zones, structural regions and roads respectively.

In previous work, we have shown how hand-crafted low-level visual features from satellite imagery like pixel intensity histograms can increase performance of a GNN on a classification task~\cite{stromann2021learning}. He \textit{et al.}~\cite{he2020roadtagger} proposed an integration of visual features from satellite imagery through CNNs as node features to a GNN. They use a generic CNN which is trained from scratch together with the GNN in an end-to-end learning framework.

\subsubsection{Transfer Learning in Remote Sensing}
Transfer learning is commonly used in remote sensing, solving challenges like scene classification~\cite{pires2020cnnTransferRS}, object detection~\cite{chen2018transferRS, zhou2020bayesian}, or semantic segmentation~\cite{cui2020semantic}. However, when transferring from a CNN pretrained on ImageNet~\cite{russakovsky2015imagenet} to different domains than natural camera image classification, a fine-tuning to the respective dataset is required. This fine-tuning may gradually span from readjusting weights in the final classification layers to readjusting even weights up to the first convolutional layers - or a careful combination of both for certain parts of the training~\cite{shin2016deep, cui2020semantic}. 

\section{Materials and Methods}

\subsection{Material}

\subsubsection{Road Network Graph}

We use a graph representation $G{=}(V,E)$ of the road network, which encodes roads as vertices $v \in V$ and connections between roads as edges $e \in E$. This representation is commonly used in GNN applications to road networks~\cite{stromann2021learning, gharaee2021graph, wu2020hrnr, liu2020graphsage, jepsen2020relational, zhang2019graph}. In a way, it resembles how humans intuitively conceptualize road networks; seeing it as connections of roads rather than connections of intersections~\cite{jilani2013multigrain}. We obtain the road network data from OSM~\cite{OpenStreetMap} through OSMnx~\cite{boeing2017osmnx}.

\subsubsection{NWPU-RESISC45}

The NWPU-RESISC45 dataset is a publicly available benchmark for REmote Sensing Image Scene Classification (RESISC), created by Northwestern Polytechnical University (NWPU)~\cite{cheng2017remote}. It contains 31,500 images, covering 45 scene classes with 700 images in each class. The dataset consists of remote sensing imagery of different spatial resolutions, and the classes represent objects of different spatial scales. We use NWPU-RESISC45 to fine-tune ResNet-models (pretrained on ImageNet) to remote sensing imagery.

\begin{figure*}[ht]
    \centering
    \includegraphics[width=\linewidth]{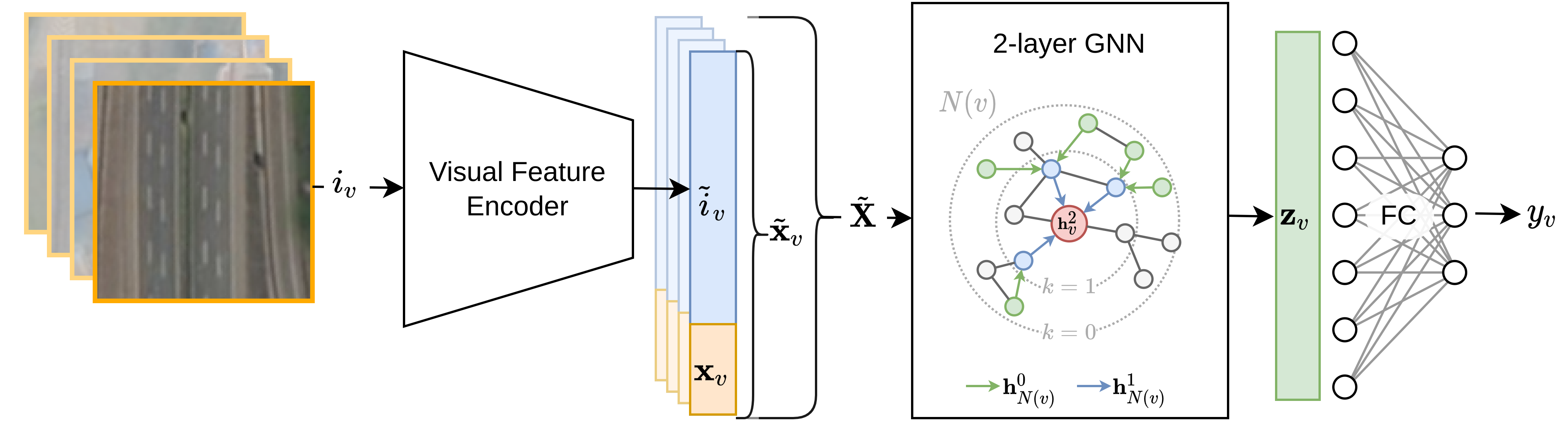}
    \caption{Overview of the dataflow in our proposed architecture.}
    \label{fig:architecture}
\end{figure*}

It should be noted, that the NWPU-RESISC45 dataset contains classes that are not explicitly related to the target task of road type classification. Instead, it contains general \textit{scenes} from land use and land cover classes (e.g., commercial area, farmland, forest, industrial area, mountain, residential area),
man-made object classes (e.g., airplane, airport, bridge, church, palace, ship), as well as landscape nature object classes (e.g., beach, cloud, island, lake, river, sea ice)~\cite{cheng2017remote}. As such, the classes have a wide variety of spatial and spectral patterns.

\subsection{Methods}
\subsubsection{GNN Layers}
\label{section:GNN}
We use graph convolutional layers (GCNs) as GNN layers. From the input graph $G{=}(V,E)$, GCNs produce node representations of the $\mathbf{h}_v^{k}$ for each node $v\in V$ at layer depth $k$. To do so, node representations of the preceding GNN-layer from neighbouring nodes $\mathbf{h}_n^{k-1}, \forall n \in N(v)$ are aggregated using an aggregation function $\textsc{Agg}^k$ and multiplied by a learnable weight matrix $\mathbf{W}^k$ before being passed through a non-linear activation function $\sigma$. 

\begin{equation}
    \label{eq:aggregation_general}
    \mathbf{h}_v^{k} = \sigma\left(\mathbf{W}^k \cdot
        \textsc{Agg}^k\left(\{\mathbf{h}_n^{(k-1)}\}\right)
    \right).
\end{equation}
At layer depth $k{=}0$ the node representation consist of the node features, such that  $\mathbf{h}_v^{0}{=}\mathbf{x}_v, \forall v \in V$. The node representations after the final layer $\mathbf{h}_v^K$ is the latent node representation - for notational convenience denoted as $\mathbf{z}_v$ -  and is used in final classification or regression layers~\cite{hamilton2017inductive}.

Depending on the architecture, the self-node representation might be concatenated to the neighbouring node representations. This is either done before the aggregation like in GCN~\cite{kipf2016semi}, such that
\begin{equation}
    \label{eq:aggregation_gcn}
    \mathbf{h}_v^{k} = \sigma\left(\mathbf{W}^k \cdot
        \textsc{Agg}^k\left(\{\mathbf{h}_v^{(k-1)}\} \oplus \{\mathbf{h}_n^{(k-1)}\}\right)
    \right),
\end{equation}
where $\oplus$ denotes concatenation, or after the aggregation like in GraphSAGE~\cite{hamilton2017inductive}, such that.

\begin{equation}
    \label{eq:aggregation_graphsage}
    \mathbf{h}_v^{k} = \sigma\left(\mathbf{W}^k \cdot \{\mathbf{h}_v^{(k-1)}\} \oplus 
        \textsc{Agg}^k\left(\{\mathbf{h}_n^{(k-1)}\}\right)
    \right).
\end{equation}

For further details on the GNN layers, we refer the interested reader to Kipf and Welling~\cite{kipf2016semi} for GCN, Hamilton \textit{et al.}~\cite{hamilton2017inductive} for GraphSAGE and to the PyTorch Geometric documentation of the implementation of these~\cite{Fey/Lenssen/2019}.



\subsection{Visual Feature Encodings for GNNs}

Our proposed architecture consists of three parts: \begin{enumerate*}[label=\roman*)]
\item a visual feature encoder (VFE),\label{item:backbone}
\item a GNN producing latent node representations and\label{item:gnn}
\item a classification head of FC layers mapping latent node representations to a road type label.\label{item:head}
\end{enumerate*}
Figure ~\ref{fig:architecture} illustrates the architecture.

The visual feature encoder or vision backbone takes image tiles as input and produces image encodings. The image encodings, concatenated with non-image node attributes, serve as input to the GNN. The GNN aggregates node features of the local neighbourhood and produces latent node representations, which in turn serve as input to a final FC layer and a soft-max activation for classification. Figure~\ref{fig:architecture} illustrates the architecture.


From the image tile $i_v$ of a node $v$, the visual feature encoder (VFE) produces the image encodings
\begin{equation}
    \tilde{i}_v = f_{\mathrm{VFE}}(i_v).
\end{equation}
The image encodings $\tilde{i}_v$ are concatenated with the other node attributes $\mathbf{x}_v$, to form the feature vector 
\begin{equation}
    \tilde{\mathbf{x}}_v = \tilde{i}_v \oplus \mathbf{x}_v,
\end{equation} 
where $\oplus$ corresponds to a vector concatenation. $\boldsymbol{\rm X}$ are the node attributes of all nodes excluding the image encodings, $\tilde{\boldsymbol{\rm X}}$ are the node features of all nodes including the image encodings, and $\boldsymbol{\rm A}$ is the adjacency matrix of the input graph $G$. 
The GNN produces a latent representation $\mathbf{z}_v$ of the self-node as described in Section~\ref{section:GNN},
such that, 
\begin{equation}
    \mathbf{z}_v = f_{\mathrm{GNN}}( \tilde{\boldsymbol{\rm X}}, \boldsymbol{\rm A}).
\end{equation}
Finally, the FC layer learns a mapping from the latent node representation $\mathbf{z}_v$ to a road type label 
\begin{equation}
    y_v=f_{\mathrm{FC}}(\mathbf{z}_v).
\end{equation}
Thus, the whole model is 
\begin{equation}
y_v=f_{\mathrm{FC}}(f_{\mathrm{GNN}}(\tilde{\boldsymbol{\rm X}}, \boldsymbol{\rm A})),
\end{equation}
where $f_{\mathrm{FC}}$ and $f_{\mathrm{GNN}}$ have learnable weights and biases and are updated using stochastic gradient descent over a categorical cross-entropy loss.

\begin{figure}[t]
    \centering
    \includegraphics[width=\linewidth]{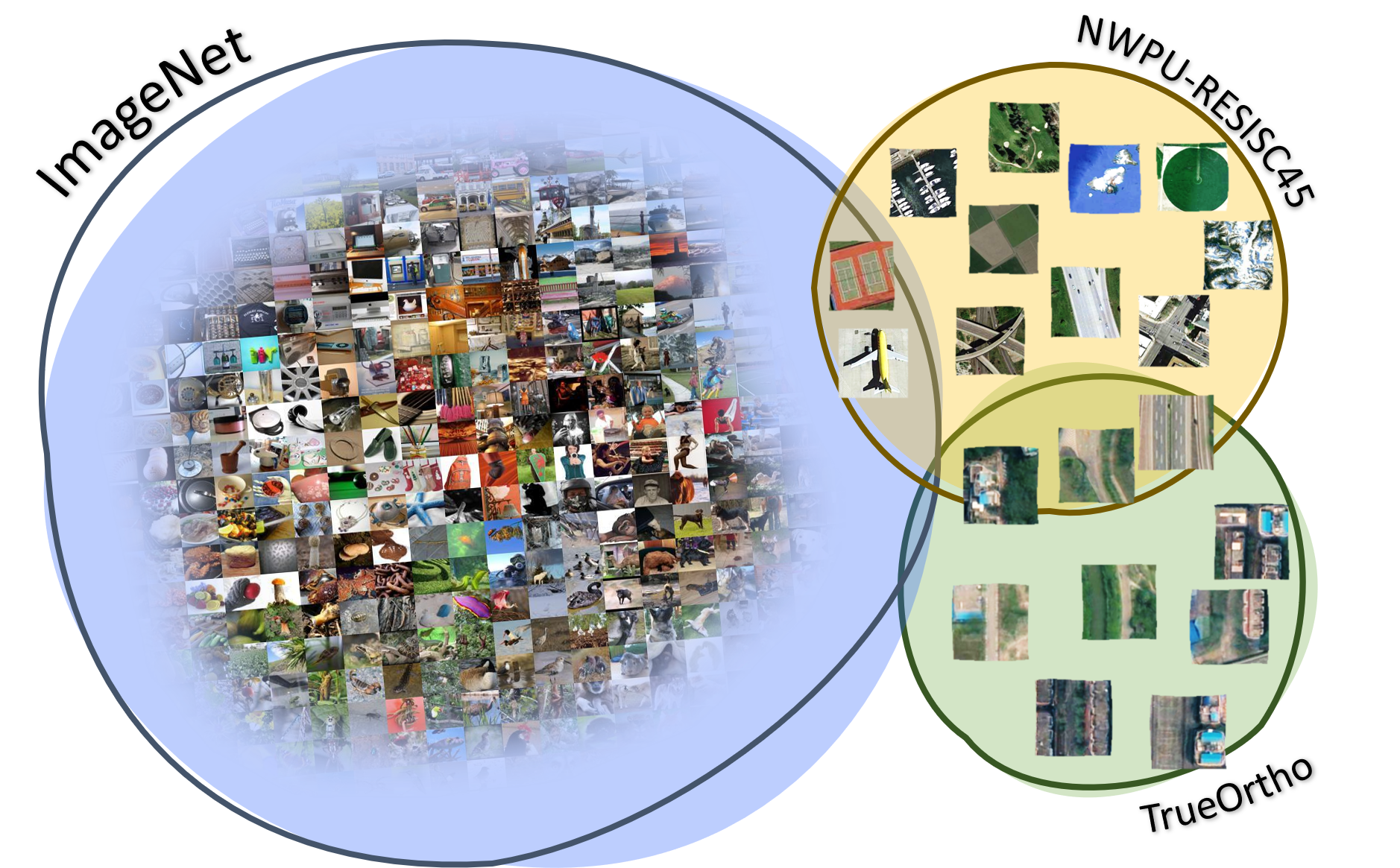}
    \caption{Conceptual view of the class membership overlap of the different datasets. ImageNet has little overlap with \textit{TrueOrtho} for road type classification. NWPU-RESISC45, though a general remote sensing dataset, has larger overlap with \textit{TrueOrtho}.}
    \label{fig:diagram}
\end{figure}

This study focusses on the visual feature encoder, and we evaluate different architectures and pretraining strategies. More specifically, we compare hand-crafted low-level visual features \cite{stromann2021learning} against visual feature encodings produced by ResNets pretrained on ImageNet. Moreover, we assess how a further fine-tuning of these ResNets to the NWPU-RESISC45 dataset improves the performance of GNN on road type classification. Conceptually, the motivation for using the NWPU-RESISC45 to adapt towards the final task of road type classification is illustrated in Figure~\ref{fig:diagram}.

All models are implemented in PyTorch Geometric~\cite{Fey/Lenssen/2019} and the code will be published.

\section{Experiments}

To demonstrate how different visual feature encodings affect the performance of a GNN on road networks, we run experiments on an exemplary machine learning task, namely node-based multi-class classifications of road type labels~\cite{stromann2021learning}.

\subsection{Dataset}

A road network from the city of Chengdu, China, is extracted from OSM~\cite{OpenStreetMap}. Through random sampling, 1842 of the nodes in $E$ are selected for validation and 1981 nodes are selected for testing. The remaining 18218 nodes are used for training.

\begin{figure}[t]
\begin{center}
   \includegraphics[width=\linewidth]{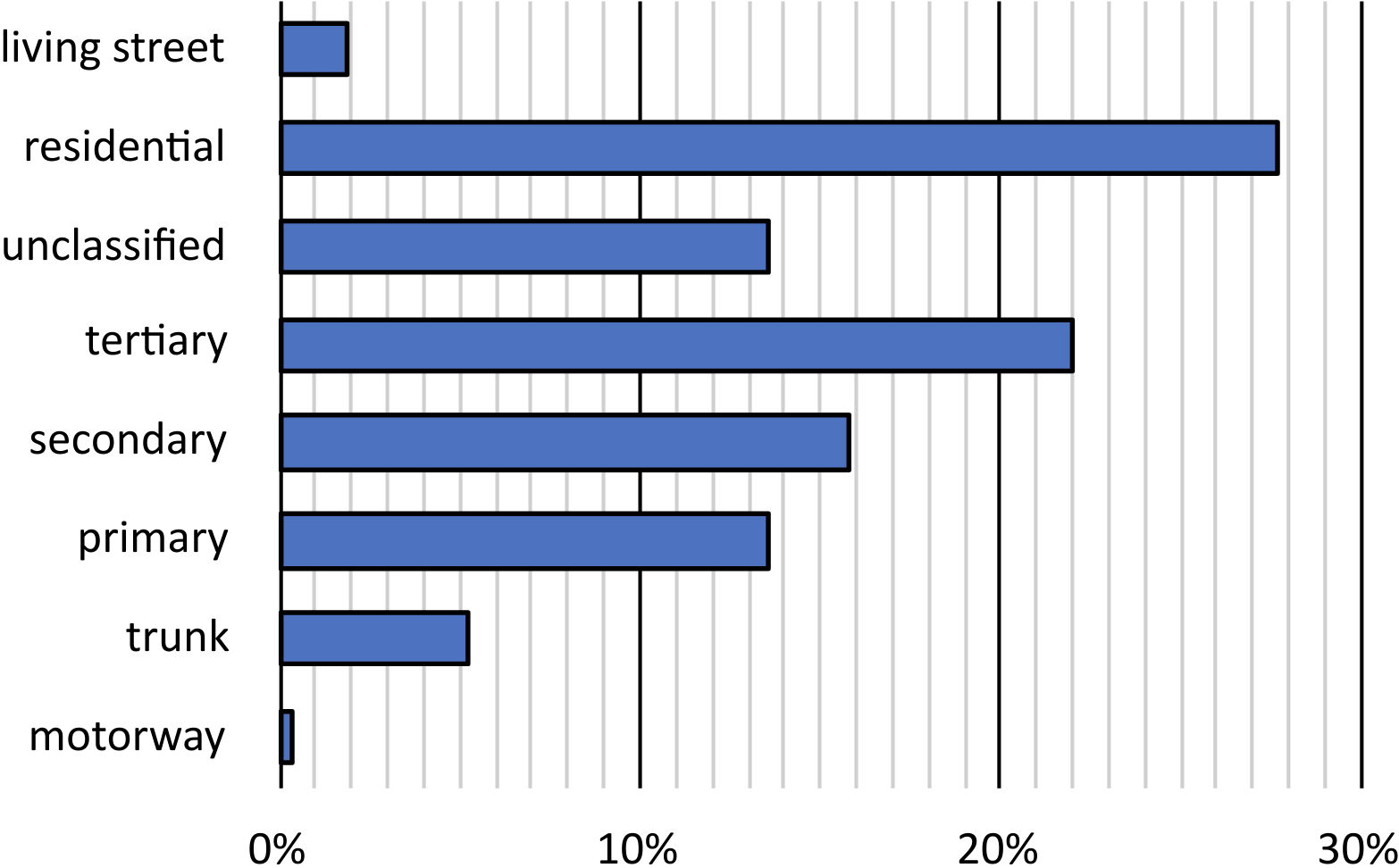}
   \caption{Class distributions in percentage. Some classes are underrepresented in the dataset.}
    \label{fig:class_dist}
\end{center}
\end{figure}

The \textit{highway} label from OSM is used as the target class label. The label describes the type of road and is an indicator for the importance of the road within the network~\cite{osm_keyhighway_nodate}. Figure \ref{fig:class_dist} illustrates the class distributions for the eight road type labels. The three classes \textit{living street}, \textit{trunk} and \textit{motorway} are underrepresented. This class imbalance makes the classification a challenging problem. 


The following attributes make up the feature vector $\mathbf{x}_v$ for each node $v\in V$ and are identical to the attributes used in our previous work~\cite{stromann2021learning}:
Geographical attributes of \textit{length}, \textit{bearing}, \textit{centroid}, and \textit{geometry}, where the latter is the road geometry resampled to a fixed number of equally-distanced points and translated by the \textit{centroid} to yield relative distances in northing and easting (meters). Binary features depict \textit{one-way}, \textit{bridge}, and \textit{tunnel}. 

These features are complemented with remote sensing data provided by Maxar Technologies~\cite{maxar2021}. Specifically, we use high-resolution analysis-ready orthorectified satellite imagery (\textit{TrueOrtho}) with a spatial resolution of \SI{0.5}{\metre}. \textit{TrueOrtho} being analysis-ready means that it is a temporal composite image in which clouds and vehicles have been removed and all buildings and structures have been rectified. \textit{TrueOrtho} contains 3-channels in RGB. 

For each node $v \in V$, we extract a 120 x 120 pixel wide image tile $i_v$ from \textit{TrueOrtho} rotated along the heading of the road.

\subsection{ResNet Fine-tuning}
\label{section:ResNet-fine-tune}
For the visual feature encoders, we firstly use ResNet-18@ImageNet and ResNet-50@ImageNet, which are the off-the-shelf ResNets pretrained on ImageNet. To compare the impact of transfer learning to remote sensing, we generate the ResNet-18@NWPU-RESISC45 and ResNet-50@NWPU-RESISC45 models by fine-tuning ResNet-18@ImageNet and ResNet-50@ImageNet to the NWPU-RESISC45 dataset. This is done, by stripping away the FC classifier head of the ResNets@ImageNet and replacing it with a similar FC with 45 output nodes. We split the NWPU-RESISC45 dataset randomly such that for each class, 600 images are used for training and 100 images are used for validation. 
We use stochastic gradient descent with momentum and a categorical cross-entropy loss to fine-tune the models. An exhaustive grid search over the learning rate $\in [0.0001, 0.001, 0.01]$ and the momentum $\in [0.9, 0.5]$ with a batch size of 512 and 100 epochs is performed to find the best models based on the validation set.

\subsection{Training}
Next we train the GNNs using the various different visual feature encoders. All models are trained in a semi-supervised, transductive setting, i.e. the input graph is complete and features of training, validation, and test nodes are reachable by the neighbourhood aggregation step of the GNN. However, only training node labels are used to compute the categorical cross-entropy loss and update the model parameters of the GNN and the FC.

The GNN consists of two layers (layer depth $K{=}2$) with a mean as aggregator function $\textsc{Agg}^k$, ReLU activations and a dropout layer with a dropout rate determined by a hyperparameter search. During training, local neighbourhoods of the target nodes are sampled using the standard GraphSAGE sampling strategy~\cite{hamilton2017inductive}. We evaluate GraphSAGE-layers and GCN-layers~\cite{hamilton2017inductive, kipf2016semi}. Finally, a single FC-layer maps the GNN encodings to road type labels.

We perform an exhaustive grid search to find the best hyperparameters for each model. Table~\ref{tab:hyperparams} shows the search ranges. Hyperparameters are selected based on the best validation set performance, but test set performances are reported. The learnable model parameters are optimized using stochastic gradient descent with a step-wise learning rate scheduler, multiplying the initial learning rate by $\gamma$ every 25 epochs. All models were trained for a maximum of 100 epochs on an NVIDIA DGX-A100, with batch sizes varying from 32 to 64 depending on the model size.

\begin{table}[t]
\centering
\caption{Hyperparameter search space.}\label{tab:hyperparams}
\begin{tabular}{| r | c |}
\hline
\textbf{Hyperparameter} & \textbf{Search Space} \\
 \hline\hline
 learning rate & $[0.5, 0.05]$\\ \hline
 lr-scheduler $\gamma$ & $[0.2, 0.5, 0.8]$ \\ \hline
 weight decay & $[0.0004, 0.0008]$  \\ \hline
 GNN dropout rate & $[0, 0.15, 0.3]$ \\ \hline
 
\end{tabular}
\end{table}

\section{Results}

\begin{figure}[htbp]
    \centering
    \includegraphics[width=\linewidth]{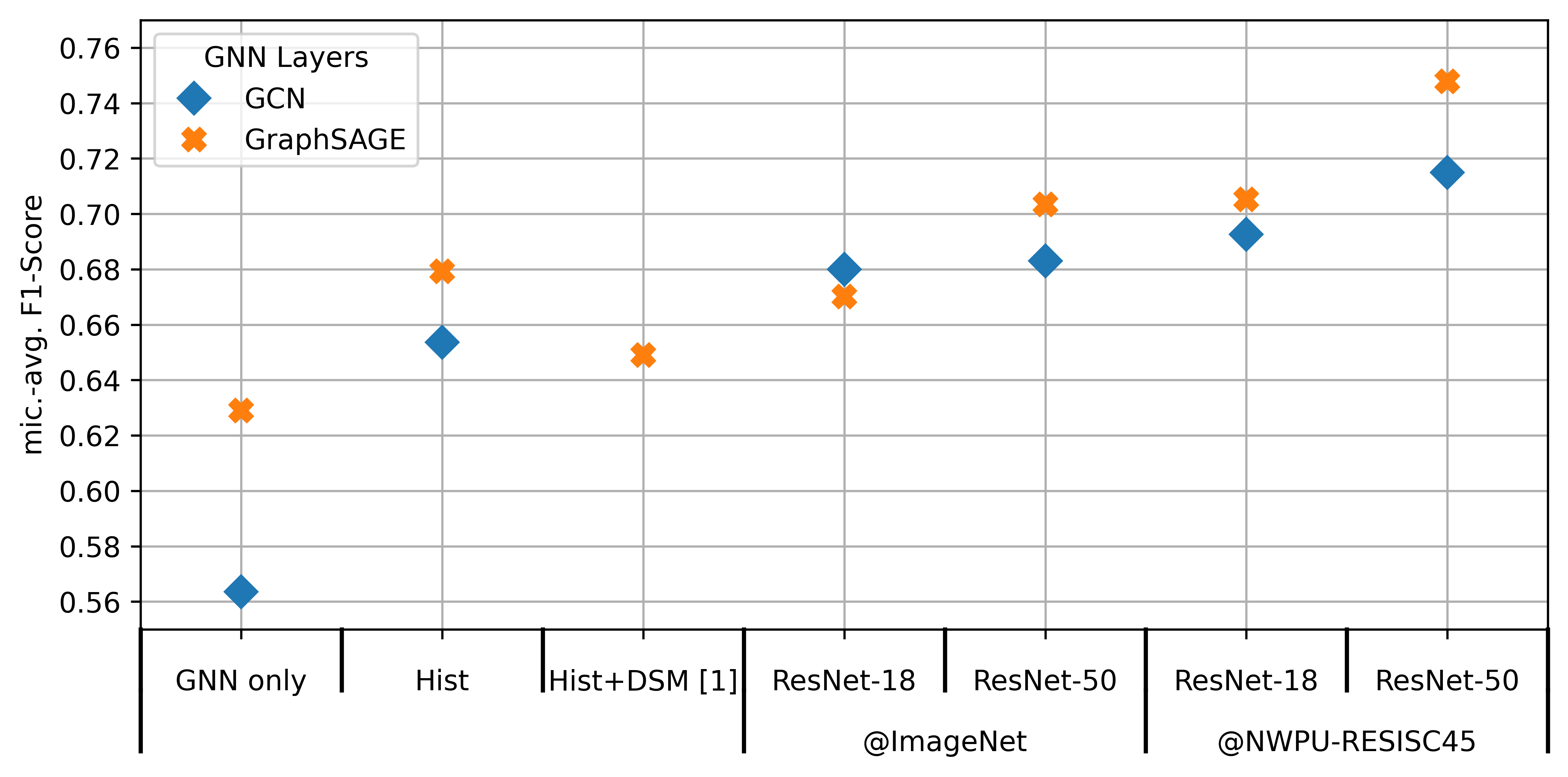}
    \caption{Micro-averaged F1-Scores for the different models, averaged over top 5 runs.}
    \label{fig:results}
\end{figure}

Figure~\ref{fig:results} shows the average micro-averaged F1-scores of the  5 top-performing models in the hyperparameter search for the 8-class road type node classification. Below is a description of the different models, in the order of appearance:
\begin{itemize}
    \item \textit{GNN only} does not use any visual feature encoder, and only node attributes $\boldsymbol{\rm X}$ are available to the GNN.
    \item \textit{Hist} means that pixel intensity histograms from the RGB satellite imagery like in~\cite{stromann2021learning} are appended to the node attributes to form the node features  $\boldsymbol{\rm \tilde{X}}$ for the GNN. 32 histogram values per channel (96 in total) are appended.
    \item \textit{Hist+DSM} has 32 additional histogram values compared to \textit{Hist} (128 in total). These stem from the digital surface model (DSM) as in~\cite{stromann2021learning}.\footnote{In~\cite{stromann2021learning} the training was done using the original GraphSAGE implementation in TensorFlow~\cite{hamilton2017inductive}, hence no result for GCN is reported.}
    \item \textit{ResNet-18} has visual feature encodings $\tilde{i}$ of size 512 produced from a pretrained ResNet-18 appended to the node features.
    \item Similarly, \textit{ResNet-50} has visual feature encodings $\tilde{i}$ of size 2048 produced from a pretrained ResNet-50 appended to the node features.
\end{itemize}

It is further indicated if the ResNets were pretrained on ImageNet (@ImageNet) or if they were pretrained on ImageNet and then fine-tuned on NWPU-RESISC45 (@NWPU-RESISC45) as described in Section~\ref{section:ResNet-fine-tune}.
The choice of GNN layers is indicated by the orange cross for GraphSAGE and the blue diamond for GCN.

It can be seen that a GNN relying only on node attributes as features performs worst, but that adding low-level visual features in the form of pixel intensity histograms increase the classification performance drastically. Visual feature encodings by ResNet-18@ImageNet increases the performance only by a negligible margin compared to the low-level visual feature, and ResNet-50@ImageNet performs just slightly better. However, a major leap in performance can be observed from ResNets that were fine-tuned on remote sensing data, especially with the larger ResNet-50@NWPU-RESISC45 which achieves 0.75 or 0.72 micro-average F1-scores for GraphSAGE or GCN architectures respectively.

\section{Discussion and Conclusion}
We have shown how visual feature encodings from state-of-the-art pretrained vision backbones like ResNets can improve the performance of GNNs on spatial graphs in a typical classification task on road networks. We also demonstrated how these visual feature encodings are superior to hand-crafted low-level visual features such as pixel intensity histograms, as previously presented in~\cite{stromann2021learning}. Moreover, we have shown that pretrained ResNets fine-tuned to the general remote sensing dataset of NWPU-RESISC45 can further boost the performance of GNNs on road type classification. 

The good performance of the ResNets@NWPU-RESISC45 suggests that the general remote sensing dataset is closer to the target image tiles of different road types extracted from \textit{TrueOrtho}. Though, there is only a small class membership overlap between NWPU-RESISC45 and \textit{TrueOrtho}, it is still larger than between ImageNet and \textit{TrueOrtho}. Figure~\ref{fig:diagram} illustrates how these datasets relate to another. By fine-tuning the visual feature encoders from ImageNet to NWPU-RESISC45, the encoders have moved closer to the target classifications in \textit{TrueOrtho}.

\section*{Acknowledgment}
We would like to gratefully acknowledge the support of Sweden’s innovation agency, Vinnova, through project iQDeep (project number 2018-02700). 

Map data copyrighted OpenStreetMap contributors and available from https://www.openstreetmap.org. 

We would like to gratefully thank Maxar Technologies for providing high-res satellite imagery.

The computations were enabled by resources provided by the Swedish National Infrastructure for Computing (SNIC), partially funded by the Swedish Research Council through grant agreement no. 2018-05973.

We used Weights \& Biases for experiment tracking and visualizations to develop insights for this paper.




%


\bibliographystyle{IEEEtran}
\bibliography{paper}

\end{document}